\documentclass[runningheads]{llncs}
\usepackage{graphicx}
\usepackage{amsmath}
\usepackage{amsfonts}
\usepackage{marvosym}
\begin{document}
\title{CLIP for Lightweight Semantic Segmentation}
\author{Ke Jin \and Wankou Yang\textsuperscript{\Letter}}

\institute{School of Automation, Southeast University, Nanjing 210096, China \email{jinke@seu.edu.cn}}

\authorrunning{Ke Jin et al.}

\titlerunning{CLIP for Lightweight Semantic Segmentation}
%
%
\maketitle              
\begin{abstract}
The large-scale pretrained model CLIP, trained on 400 million image-text pairs, offers a promising paradigm for tackling vision tasks, albeit at the image level. Later works, such as DenseCLIP and LSeg, extend this paradigm to dense prediction, including semantic segmentation, and have achieved excellent results. However, the above methods either rely on CLIP-pretrained visual backbones or use none-pretrained but heavy backbones such as Swin, while  falling ineffective when applied to lightweight backbones. The reason for this is that the lightweitht networks, feature extraction ability of which are relatively limited, meet difficulty embedding the image feature aligned with text embeddings perfectly. In this work, we present a new feature fusion module which tackles this problem and enables language-guided paradigm to be applied to lightweight networks. Specifically, the module is a parallel design of CNN and transformer with a two-way bridge in between, where CNN extracts spatial information and visual context of the feature map from the image encoder, and the transformer propagates text embeddings from the text encoder forward. The core of the module is the bidirectional fusion of visual and text feature across the bridge which prompts their proximity and alignment in embedding space. The module is model-agnostic, which can not only make language-guided lightweight semantic segmentation practical, but also fully exploit the pretrained knowledge of language priors and achieve better performance than previous SOTA work, such as DenseCLIP, whatever the vision backbone is. Extensive experiments have been conducted to demonstrate the superiority of our method.

\end{abstract}
\section{Introduction}
In recent years, the CV community has made a lot of efforts to apply the achievements of NLP to the processing of visual tasks, and CLIP~\cite{ref1} is one of the most successful language-guided methods. In order to leverage a much broader source of supervision, namely the unlimited raw text on the Internet, and embrace the generality and usability that traditional supervised learning does not have, the authors of CLIP perform comparative learning training on 400 million image-text pairs to obtain a pair of image-text encoders. They demonstrate that the simple pretraining task of predicting which caption goes with which image is an efficient and scalable way to learn state-of-the-art image and text representations aligned in embedding space. After pretraining, natural language is used to reference learned visual concepts (or describe new ones), enabling zero-shot transfer of the model to downstream tasks such as OCR, action recognition in videos, geo-localization and object classification, which can basically be classified as image-level prediction tasks. 

Shortly after CLIP was proposed, the problem of transferring the knowledge learned from image-text pairs to more complex dense prediction tasks, such as object detection and semantic segmentation, is visited by the community quickly~\cite{ref2,ref3}. The core of the problem is that, compared to image-level tasks such as object classification, pixel-level tasks not only require the ability to distinguish the concepts represented by the image, but also need to use spatial information to locate these concepts corresponding to pixels.

To solve the above-mentioned difficulty, scholars mostly convert the original image-text matching problem in CLIP to a pixel-text matching problem. Because they observed an interesting property of the feature map before the final pooling of CLIP's visual encoder, that it not only preserves spatial information, but also aligns with the text embeddings to some extent but not perfectly. Then they construct pixel-text score maps through inner product or other methods, to guide the learning of dense prediction models explicitly.

However, we notice that these methods rely on CLIP vision pre-training or heavy vision encoders trained on massive data, such Swin~\cite{ref4}, and fall ineffective when applied to lightweight backbones. For an example, when using MobileNet~\cite{ref5} trained on ADE20K dataset~\cite{ref6} as a vision encoder, image feature and text feature can't be well aligned in embedding space, since the feature extraction ability of lightweight backbone is relatively limited compared to large models.

In this work, we propose a new feature fusion module which enables language-guided paradigm to be applied to lightweight network. The module takes inspiration from Mobile-Former~\cite{ref7} and is designed as a a parallel architecture of CNN and transformer with a two-way bridge in between. On the one hand, the CNN takes the feature map from the vision encoder as input and stacks inverted bottleneck blocks. It leverages the efficient depthwise and pointwise convolution to extract spatial information and visual context of the feature map. On the other hand, the transformer takes the text embeddings from text encoder as input and stacks multi-head attention and feed-forward blocks. 

The CNN and transformer in our module communicate through a two-way bridge to fuse the visual and text embedding and make them aligned by performing a lightweight cross attention we propose. The bridge, which is set at every bottleneck of CNN, feeds the image feature to text tokens in the transformer and introduces the text information to every pixel in the CNN reversely.

Extensive experiments demonstrate our method can greatly improve the segmentation performance with inference time slightly increasing which can be viewed as an an acceptable compromise. The results of comparative experiments of multiple groups prove our following claims:
\begin{enumerate}
    \item [(1)] The fusion module we propose can solve the difficulty that the language-guided paradigm cannot be well applied to lightweight visual backbones.
    \item [(2)] The fusion module we propose is model-agnostic. It can fully exploit the pretrained knowledge of language priors and achieve better performance than previous SOTA work, DenseCLIP, even based on CLIP-pretrained models which DenseCLIP is designed for.
\end{enumerate}

\begin{figure}
    \centering
    \includegraphics[width=\textwidth]{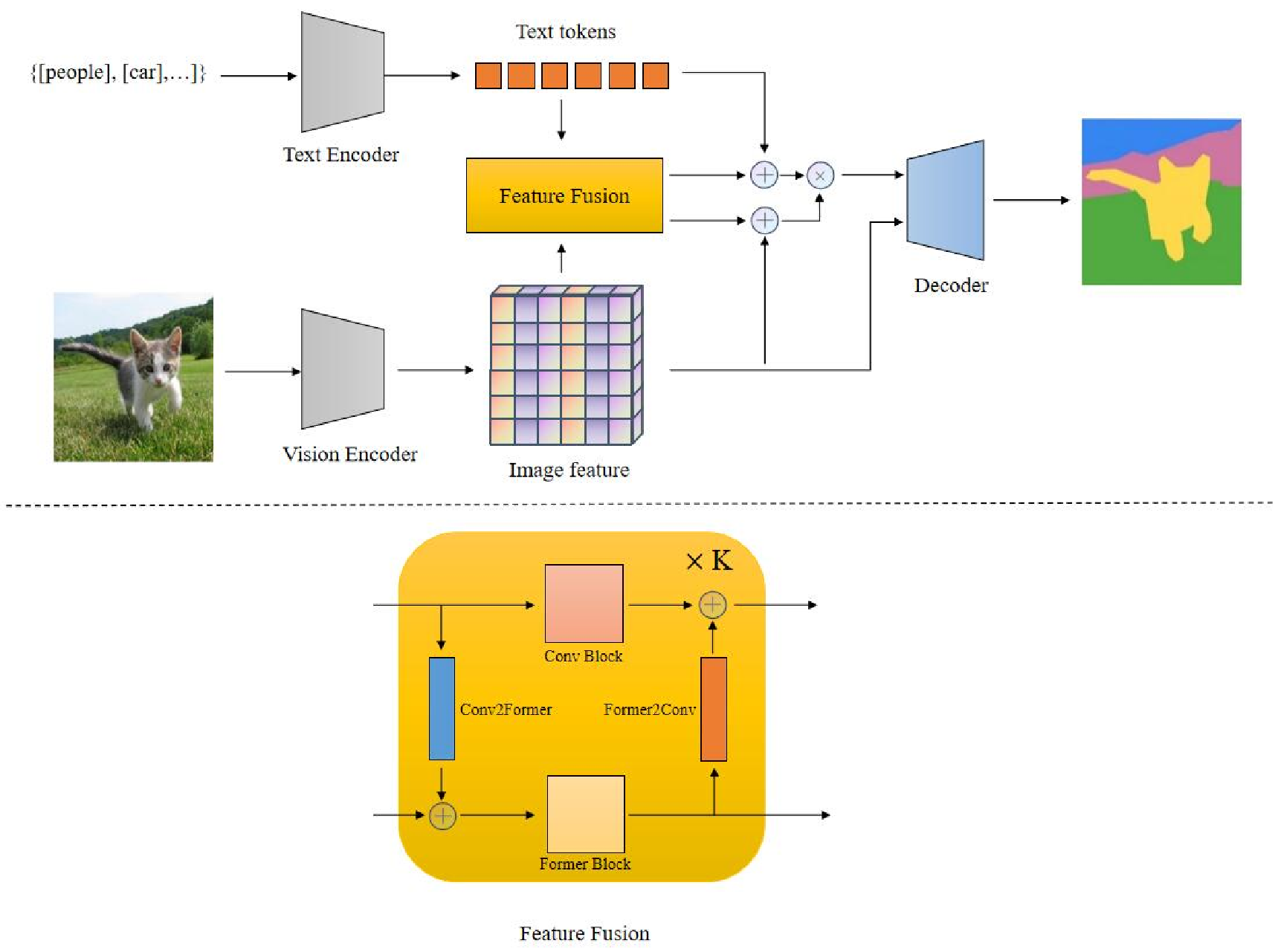}
    \caption{The overall framework of our method and the pipeline of language-guided segmentation is demonstrated.  }
    \label{fig1}
    \vspace{-0.5cm}
\end{figure}

\section{Related Work}
\subsection{Lightweight Visual Encoder}
The deep neural network has changed the face of the field of visual recognition in a subversive way in the past ten years, but its high requirements for hardware computing performance have become a major obstacle to its application in production and life. Therefore, the research of lightweight network has become the focus of people's attention, which bridges the gap between academic research and industrial application.

MobileNetV2~\cite{ref5} is based on an inverted residual structure where the shortcut connections are between the thin bottleneck layers and propose lightweight depthwise convolution to save computational cost. ShuffleNet~\cite{ref15} utilizes two proposed operations, pointwise group convolution and channel shuffle, to greatly reduce computation cost while maintaining accuracy. Combining the previous two works, Xception~\cite{ref13} proposes the depthwise separable convolution operation (a depthwise convolution followed by a pointwise convolution).

With the popularity of vision transformers, how to design lightweight transformers has also been visited frequently by the community. Swin~\cite{ref4} and following works propose window-based attention such that the receptive field is constrained to a pre-defined window size, which also inspires subsequent work to refine attention patterns~\cite{ref16,ref17}. Another track is to combine lightweight CNN and attention mechanism to form a hybrid architecture, such as MobileVit~\cite{ref18} and EfficientFormer~\cite{ref14}.

\subsection{Language-guided Recognition}Language-driven recognition is an active area of research. Common tasks in this space include visual question answering~\cite{ref21}, image captioning~\cite{ref20}, and image-text retrieval~\cite{ref19}. CLIP~\cite{ref1} uses contrastive learning together with high-capacity language models and visual feature encoders to synthesize extremely robust models for zero-shot image classification. Later works~\cite{ref2,ref3} convert the original image-text matching problem in CLIP to a pixel-text matching problem and construct pixel-text score maps to guide the inference of dense prediction. Lseg~\cite{ref3} gets highly competitive zero-shot performance compared to existing zero- and few-shot semantic segmentation methods, while DenseCLIP~\cite{ref2} designs a more sophisticated framework and achieves SOTA.

\section{Proposed Method}
\subsection{Language-guided Semantic Segmentation Framework}
The pipeline of the language-guided semantic segmentation framework is shown in Fig.~\ref{fig1}. Before each segmentation, text prompts from the template “a photo of a [CLS].” with K class names are used as the input of the text encoder(CLIP-pretrained) to obtain text embeddings $T\in \mathbb{R}^{K\times C}$. After training, these embeddings are stored in memory for use in forward inference of the model, which can reduce the overhead brought by text encoder. On the other hand, the image encoder extract a language-compatible feature map $I\in \mathbb{R}^{H\times W\times C}$ from the input picture. Then $T$ and $I$ are both fed into the feature fusion module and realigned in the embedding space by this module.
\begin{equation}
    [T',I']= \mathbf{\text{Conv-Former}}([T,I])+[T,I]
\end{equation}
After that, we correlate the visual and text embeddings by the inner product, creating a tensor of size $H\times W\times K$, defined as
\begin{equation}
    S=I'\cdot T'^{T}
\end{equation}
The tensor $S$ is the score map we need and it characterizes the results of pixel-text matching. We can view the score map as segmentation results with a lower resolution and concatenate it to the last feature map to explicitly incorporate language priors, i.e., $X=[I',S]\in \mathbb{R}^{H\times W\times (C+K)}$. The modified feature map now can be directly used as usual in segmentation, followed by a popular decoder like semantic FPN.

\begin{figure}
    \centering
    \includegraphics[width=0.75\textwidth]{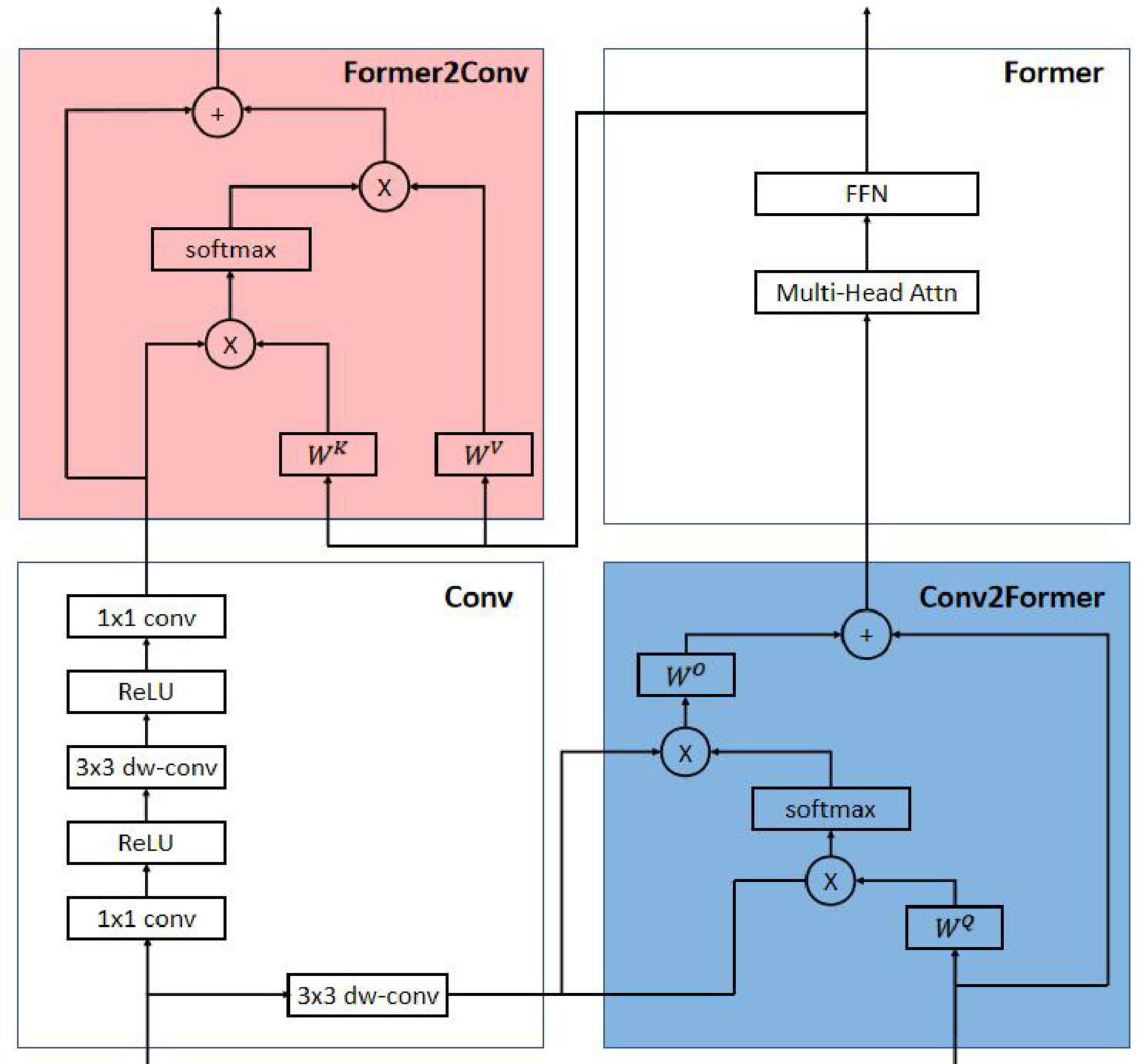}
    \caption{Conv-Former consists of four blocks, namely $Conv$, $Former$, $Conv2Former$ and $Former2Conv$, forming a parallel structure of CNN and transformer.}
    \label{fig2}
\end{figure}

\subsection{Structure of the Fusion Module}
As shown in Fig.~\ref{fig1}, the fusion module is formed by repeated stacking of the same component, which we named Conv-Former. Conv-Former consists of four blocks, namely $Conv$, $Former$, $Conv2Former$ and $Former2Conv$, forming a parallel structure of CNN and transformer. The original image feature map $I\in \mathbb{R}^{H\times W\times C}$, is fed into the first $Conv$, while the first $Former$ takes the corresponding text embedding $T\in \mathbb{R}^{K\times C}$ as input. 

Efficient convolutional layers in the $Conv$, such as depthwise and pointwise convolution, extract the spatial information of the image feature map. $Former$ leverage MultiHead-Attention operation to propagates the text embeddings forward. The bidirectional bridge, namely $Conv2Former$ and $Former2Conv$, achieves information interaction and feature fusion between the two sides through a lightweight cross attention.

\subsection{Design of Conv-Former}
\subsubsection{Bidirectional Bridge}
\paragraph{Overall:} $Conv$ and $Former$ communicate with each other through this two-way bridge. The two direction are realized by $Conv2Former$ and $Former2Conv$ respectively. A lightweight cross attention operation is proposed to model the process of feature fusion, where the projections $(W^{Q},W^{K},W^{V})$ are removed from $Conv$ side to save computations, but kept at $Former$ side. $Conv2Fomer$ fuses image feature to text tokens before both embeddings enter $Conv$ and $Former$, while $Former2Conv$ does the reverse operation after $Conv$ and $Former$ output them. In this back and forth process, the visual and the text embeddings get closer to each other in the embedding space.

\paragraph{Conv2Former:} As shown in Fig.~\ref{fig2}, the input feature map is first down-sampled to reduce computational cost by a convolutional layer with stride and kernel size of 3. The effect of convolution can be regarded as aggregating each $3\times 3\times C$ patch into $1\times C$ tokens, i.e., $I\in \mathbb{R}^{H\times W\times C}\longrightarrow \mathbf{x} \in \mathbb{R}^{\frac{H}{3} \frac{W}{3}\times C}$. The lightweight cross attention from image feature to text embeddings is computed as:
\begin{equation}
    A_{\mathbf{x}\to I}=\text{MHSA}\left(IW^{Q},\mathbf{x},\mathbf{x}\right)\cdot W^{O}\in \mathbb{R}^{K\times C}
\end{equation}
The $W^{Q}$ is the query projection matrix and $W^{O}$ is the output projection matrix which is used to combine multiple heads together. MHSA refers to the standard MultiHead-Attention function over query $Q$, $K$, and $V$ as $softmax\left(\frac{QK^{T}}{\sqrt{d_{k}}}\right)\cdot V$. Specifically, $softmax\left(\frac{IW^{Q}(\mathbf{x})^{T}}{\sqrt{d_{k}}}\right)$ denotes the correlation matrix between each image patch and label token. Then $\left(softmax\left(\frac{IW^{Q}(\mathbf{x})^{T}}{\sqrt{d_{k}}}\right)\cdot V\right)_{i}$ is the visual context information to be injected into the $i^{th}$ label embedding, which is the weighted sum of each patch feature, with $\left(softmax\left(\frac{IW^{Q}(\mathbf{x})^{T}}{\sqrt{d_{k}}}\right)\right)_{i}$ serving as the weights.
\paragraph{Former2Conv:} This block does the reverse operation of $ConvFormer$. Similarly, the cross attention from text embeddings to feature map is computed as: 
\begin{equation}
    A_{I\to \mathbf{x}}=\text{MHSA}\left(\mathbf{x},IW^{K},IW^{V}\right) \in \mathbb{R}^{H\times W\times C}
\end{equation}
In this function, $\left(softmax\left(\frac{\mathbf{x}(IW^{K})^{T}}{\sqrt{d_{k}}}\right)\cdot IW^{V}\right)_{ij}$ denotes the text context information to be injected into each pixel.

\vspace{4 ex}

\subsubsection{Conv and Former Block}
\paragraph{Conv} takes the feature map X as input and its output is taken as the input for $Former2Conv$. It uses a typical inverted bottleneck block and depthwise convolution in~\cite{ref5}. As shown in Fig.~\ref{fig3}, the kernel size of depthwise convolution is 3×3 for all blocks.
\paragraph{Former} is a standard transformer block including a Multi-Head Self-Attention (MHSA) and a feed-forward network (FFN). Expansion ratio 2 (instead of 4) is used in FFN. We follow~\cite{ref11} to use post layer normalization. $Former$ is processed between $Conv2Former$ and $Former2Conv$.

\section{Experiments}
To evaluate the effectiveness of our method, extensive experiments have been conducted on ADE20K and Cityscape~\cite{ref12}. Some representative vanilla methods and SOTA method DenseCLIP are taken as baseline. We hope to verify the following two points through experiments:
\begin{enumerate}
    \item [(1)] The fusion module we propose can solve the difficulty that the language-guided paradigm cannot be well applied to lightweight visual backbones.
    \item [(2)] The fusion module we propose is model-agnostic. It can fully exploit the pretrained knowledge of language priors and achieve better performance than DenseCLIP, even based on CLIP-pretrained models which DenseCLIP is designed for.
\end{enumerate}
\subsection{Set up and Implementation}
 Following common practice~\cite{ref8,ref9}, we report the mIoU on the validation set of ADE20K and Cityscape. Since our method is designed for lightweight semantic segmentation, we also include the GFLOPs to evaluate the computation cost.
 
 CLIP pretrained text encoder is used to generate text embeddings and We fix the text encoder during training to preserve the natural language knowledge.
 For fair comparisons, we take the Semantic FPN as the decoder.

 The setting of the learning rate is special that the learning rate of the image encoder needs to be set to $\frac{1}{10}$ of other parts to preserve the knowledge in it, regardless of whether the backbone is CLIP-pretrained or ImageNet-pretrained. AdamW~\cite{ref10} is used instead of the vanilla SGD when the vision backbone is a transformer, following previous work. 
\subsection{for Lightweight Visual Backbone}
We selected three representative lightweight visual backbone as the experimental objects, namely MobileNetV2, Xception~\cite{ref13}, and EfficientFomer~\cite{ref14}. MobileNetV2 and Xception are both typical convolutional neutral network (CNN), while EfficientFormer combines lightweight CNN and attention mechanism to form a hybrid architecture. The experiment is conducted on the ADE20K, and the results are presented in Table~\ref{tab1}.

From the experimental results, it can be seen that on the challenging ADE20K, DenseCLIP has a weak effect on performance improvement, and even get lower prediction accuracy than the vanilla method (only Semantic FPN) when it comes to MobileNetV2 and Xception. In contrast, our method achieve excellent performance that it is $+7.1\%$, $+4.8\%$ and $+3.8\%$ mIoU higher than the vanilla method (only Semantic FPN) and $+9.9\%$, $+6.5\%$ and $+3.4\%$ mIoU higher than DenseClip, with GFLOPs slightly increasing which can be viewed as an an acceptable compromise.

DenseCLIP's poor performance proves the necessity of the feature fusion module, while  excellent performance of our method verifies the effectiveness of the module.

\begin{table}\centering
\caption{Semantic segmentation results on ADE20K: We report the results of three methods respectively. By comparing DenseCLIP and our method, the necessity and effectiveness of the feature fusion module we propose are justified.}
\label{tab1}
\setlength{\tabcolsep}{4mm}
{
\begin{tabular}{|c|c|c|c|}
\hline
Backbone &  Method & mIoU & GFLOPs\\
\hline
            & Semantic FPN & 25.1 & 32.7 \\
MobileNetV2 & DenseCLIP + Semantic FPN & 22.3 & 44.5\\
            & Our Method + Semantic FPN & \textbf{32.2} & 41.3 \\
\hline
            & Semantic FPN & 35.5 & 40.2  \\
Xception    & DenseCLIP + Semantic FPN & 33.8 & 51.6\\
            & Our Method + Semantic FPN & \textbf{40.3} & 47.5 \\
\hline
            & Semantic FPN & 42.1 & 48.9 \\
EfficientFormer & DenseCLIP + Semantic FPN & 42.5 & 59.4\\
            & Our Method + Semantic FPN & \textbf{45.9} & 53.9 \\
\hline
\end{tabular}
}
\end{table}

\begin{table}
\caption{Semantic segmentation results on \textbf{ADE20K} and \textbf{Cityscapes}}
\label{tab2}
\resizebox{\textwidth}{!}
{
\begin{tabular}{|c|c|c|c|c|}
\hline
Backbone &  Method & Pretrained & mIoU(ADE20K) & mIoU(Cityscapes)\\
\hline
            & Semantic FPN & ImageNet & 38.6 & 74.5 \\
ResNet-50   & DenseCLIP + Semantic FPN & CLIP & 43.5 & 75.9\\
            & Our Method + Semantic FPN & CLIP & \textbf{44.9} & \textbf{76.3} \\
\hline
            & Semantic FPN & ImageNet & 40.4 & 75.8  \\
ResNet-101  & DenseCLIP + Semantic FPN & CLIP & 45.1 & 77.1\\
            & Our Method + Semantic FPN & CLIP & \textbf{46.7} & \textbf{77.5}\\
\hline
            & Semantic FPN & ImageNet & 48.3 & 80.5\\ 
VIT-B       & DenseCLIP + Semantic FPN & CLIP & 50.6 & 81.1\\
            & Our Method + Semantic FPN & CLIP & \textbf{51.2} & \textbf{81.3}\\
\hline
            & UperNet & ImageNet & 44.5 & 79.9\\ 
Swin-T       & DenseCLIP + UperNet & ImageNet & 45.4 & 80.2\\
            & Our Method + UperNet & ImageNet & \textbf{45.9} & \textbf{80.3}\\
\hline
\end{tabular}
}
\end{table}

\subsection{for any Visual Backbone}
In the above, we claim that the fusion module we propose is not only effective for lightweight models and has outstanding generalization ability. Therefore, we selected a representative heavy model Swin-T for experiments. Moreover, we also selected three CLIP-pretrained visual backbones, which DenseCLIP is designed for, to verify the fact that our method can fully exploit the pretrained knowledge of language priors and achieve better performance than DenseCLIP.

Table~\ref{tab2} shows the results on ADE20K and Cityscapes respectively. Our Method achieves $+1.4\%$, $+1.6\%$, $+0.6\%$ and $+0.5\%$ higher mIoU than DenseCLIP on ADE20K, and on Cityscapes the increase is $+0.4\%$, $+0.4\%$, $+0.2\%$ and $+0.1\%$. This performance meets our expectations and verifies our claims.

\subsection{Ablation Study}
In this part, we want to further study the components of Conv-Fomer and demonstrate the effect of the lightweight cross attention operation. In addition, how many Conv-Formers need to be stacked repeatedly in the fusion module is also worth exploring. Therefore, we conduct ablation experiments for the above two problems.
\begin{figure}
    \centering
    \includegraphics[width=0.75\textwidth]{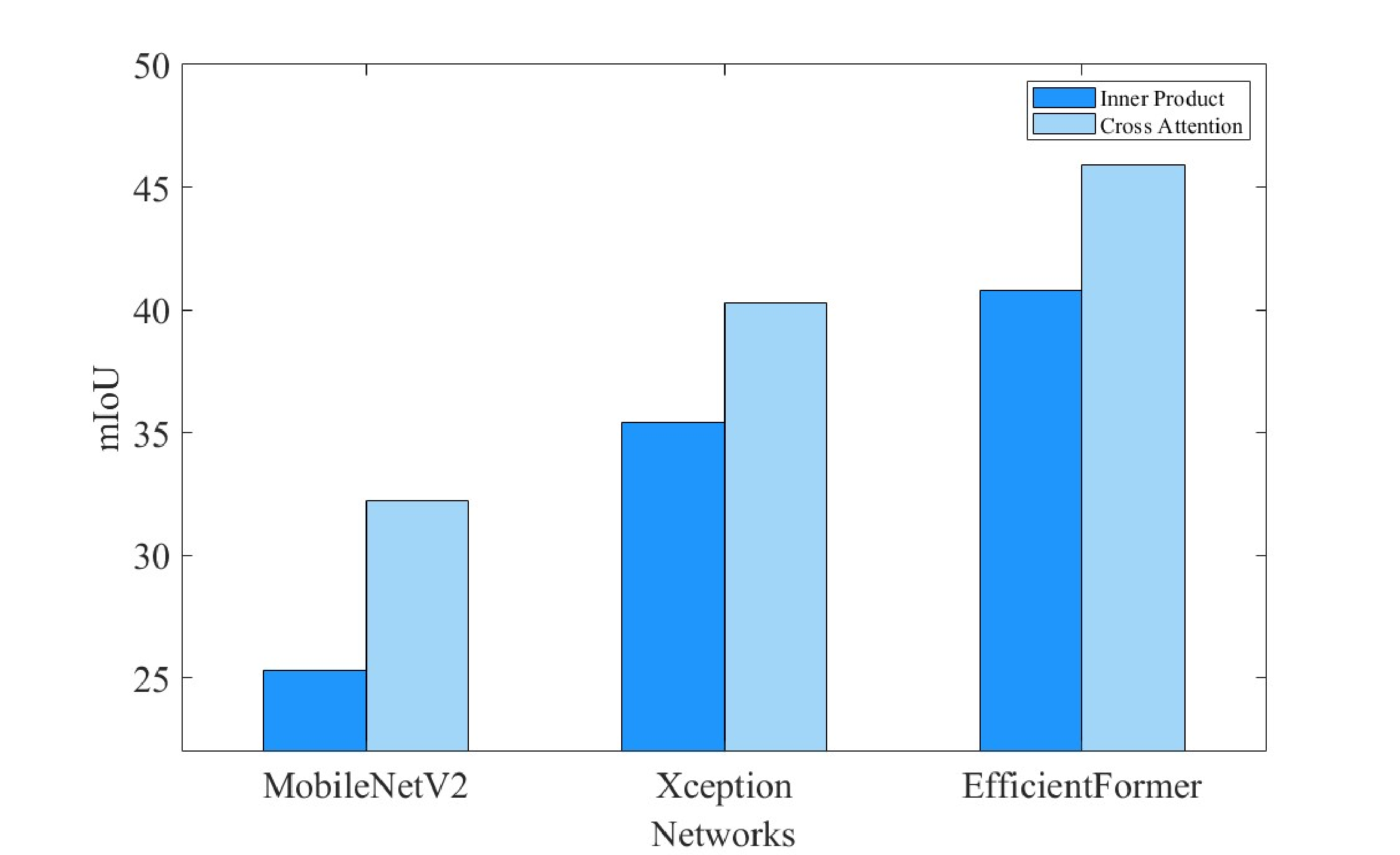}
    \caption{the comparative experimental results on three lightweight networks}
    \label{fig3}
\end{figure}
\subsubsection{Effect of the Cross Attention}
In order to dispel doubts about whether cross-attention is really necessary, we replace it with an inner product operation, which can also calculate the similarity between two embeddings. Fig.~\ref{fig3} shows the comparative experimental results on three lightweight networks, namely MobileNetV2, Xception and EfficientFomer. The performance gap exhibited in the figure illustrates the necessity of cross-attention.
\subsubsection{Number of stacked Conv-Former}
\begin{figure}
    \centering
    \includegraphics[width=0.75\textwidth]{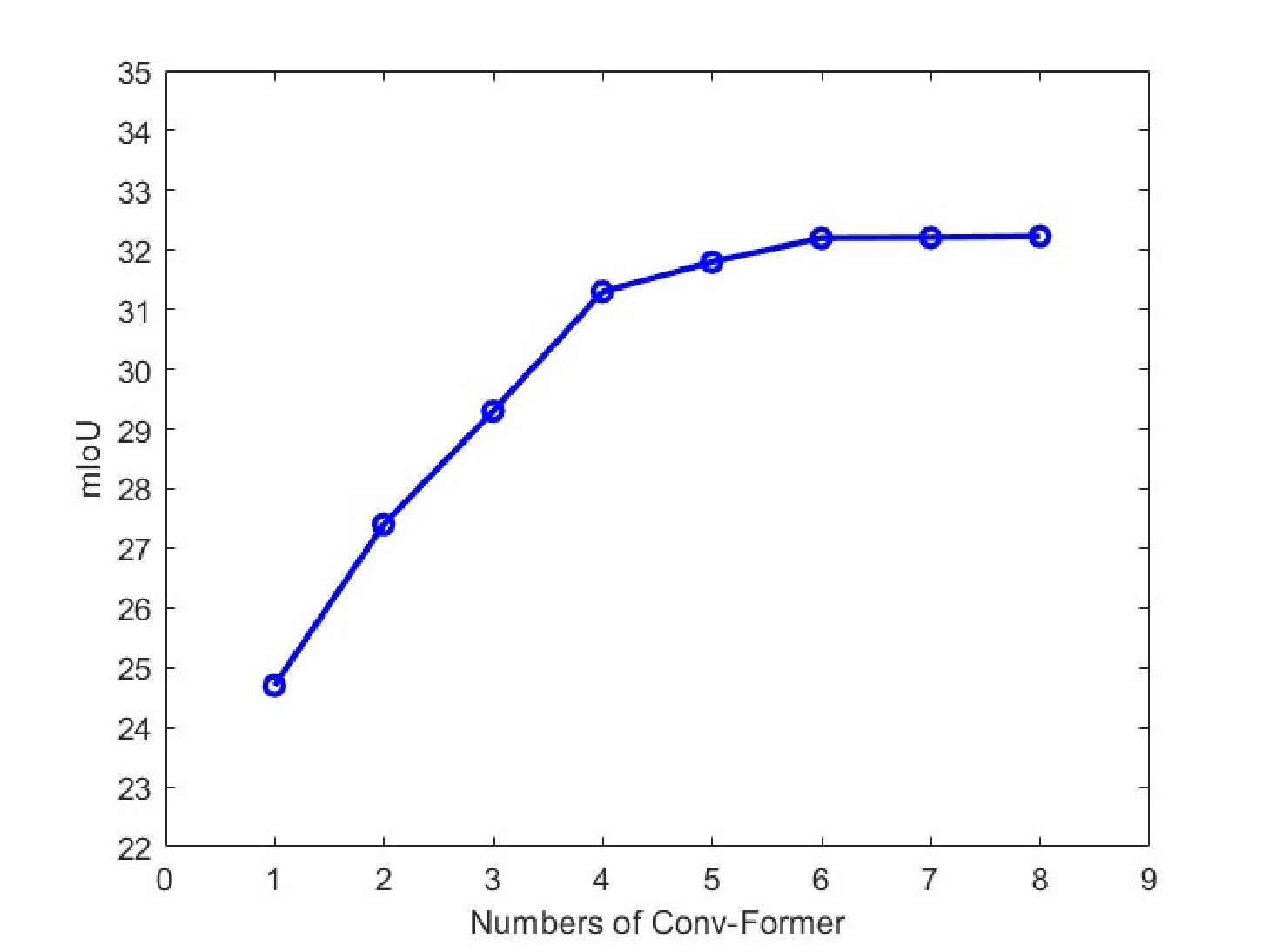}
    \caption{model performance curve with the number of blocks}
    \label{fig4}
\end{figure}

In order to explore how many Conv-Formers are most suitable for stacking in the feature fusion module, we increase the number one by one and conduct experiments separately. This research is based on MobileNetV2, whose performance is most significantly improved by the feature fusion module. 
As shown in Fig.~\ref{fig4}, stacking 6 times is the best choice, since when the number is less than 6, the performance is not optimal, and when the number is greater than 6, the performance tends to be saturated and wastes computing overhead.

\section{Conclusion}
In this paper, we propose a feature fusion module to make language-guided lightweight semantic segmentation practical. The module is model--agnostic and achieve better performence than previous SOTA work. We conducted extensive experiments to demonstrate the superiority of our method.
%
%
%
%

\end{document}